\documentclass[10pt,twocolumn,twoside]{IEEEtran}
\usepackage{cite}
\usepackage[cmex10]{amsmath}
\usepackage{amssymb}
\usepackage{graphicx}
\usepackage{cases}
\usepackage{mdwmath}
\usepackage{mdwtab}
\usepackage{array}
\usepackage{url}
\usepackage[ruled,vlined]{algorithm2e}
\usepackage{multirow}
\usepackage{booktabs}
\usepackage{threeparttable}
\usepackage{color}

\def\ie{\emph{i.e.}}
\def\eg{\emph{e.g.}}
\def\etal{{\em et al.}}

\newcommand{\figref}[1]{Fig. \ref{#1}}
\newcommand{\tabref}[1]{Tab. \ref{#1}}

\newcommand{\myPara}[1]{\vspace{.05in}\noindent\textbf{#1}}

\newcommand{\bl}[1]{\textbf{#1}}
\newcommand{\mc}[1]{\mathcal{#1}}
\newcommand{\mb}[1]{\mathbb{#1}}
\graphicspath{{./figs/}}
\usepackage{color}
\begin{document}

%\title{Learning a crowdsourced Metric for Measuring  \\ Perceptual Similarity of Saliency Maps}
\title{Learning a Saliency Evaluation Metric Using Crowdsourced Perceptual Judgments}
\author{Changqun~Xia,~Jia~Li,~Jinming Su~and~Ali Borji
\thanks{C. Xia, J. Li and J. Su are with the State Key Laboratory of Virtual Reality Technology and Systems, School of Computer Science and Engineering, Beihang University, Beijing, China.}
\thanks{J. Li is also with the International Research Institute for Multidisciplinary Science, Beihang University, Beijing, China.}
\thanks{A. Borji is with the Center for Research in Computer Vision, Computer Science Department, University of Central Florida, Orlando, Florida.}
\thanks{An earlier version of this work has been published in ICCV~\cite{li2015metric}.}
\thanks{Correspondence author: Jia Li. E-mail: jiali@buaa.edu.cn.}
%\thanks{Manuscript received Sept. 29, 2014}
}

\markboth{~}%
{Li \MakeLowercase{\textit{et al.}}}

\maketitle

\begin{abstract}
In the area of human fixation prediction, dozens of computational saliency models are proposed to reveal certain saliency characteristics under different assumptions and definitions. As a result, saliency model benchmarking often requires several evaluation metrics to simultaneously assess saliency models from multiple perspectives. However, most computational metrics are not designed to directly measure the perceptual similarity of saliency maps so that the evaluation results may be sometimes inconsistent with the subjective impression. To address this problem, this paper first conducts extensive subjective tests to find out how the visual similarities between saliency maps are perceived by humans. Based on the crowdsourced data collected in these tests, we conclude several key factors in assessing saliency maps and quantize the performance of existing metrics. Inspired by these factors, we propose to learn a saliency evaluation metric based on a two-stream convolutional neural network using crowdsourced perceptual judgements. Specifically, the relative saliency score of each pair from the crowdsourced data is utilized to regularize the network during the training process. By capturing the key factors shared by various subjects in comparing saliency maps, the learned metric better aligns with human perception of saliency maps, making it a good complement to the existing metrics. Experimental results validate that the learned metric can be generalized to the comparisons of saliency maps from new images, new datasets, new models and synthetic data. Due to the effectiveness of the learned metric, it also can be used to facilitate the development of new models for fixation prediction.

\end{abstract}


\section{Introduction}

\IEEEPARstart{I}{n} the past decades, hundreds of visual saliency models have been proposed for fixation prediction. Typically, these models are designed to reveal certain characteristics of visual saliency under different assumptions and definitions, such as Local Irregularity~\cite{li2016measuring,hu2005adaptive,riche2012rare}, Global Rarity \cite{kruthiventi2016saliency,borji2012exploiting,goferman2010context}, Temporal Surprise~\cite{liu2016saliency,kim2014spatiotemporal,mahadevan2010spatiotemporal,li2010probabilistic}, Entropy Maximization~\cite{zhang2016saliency,kummerer2014howclose,hou2013visual,wang2010measuring,bruce2005saliency} and Center-bias~\cite{borji2016reconciling,li2014visual,tseng2009quantifying}. Consequently, to conduct a comprehensive and fair comparison, it is necessary to evaluate saliency models from multiple perspectives.

\begin{figure}[t]
\centering
\includegraphics[width=1.00\columnwidth]{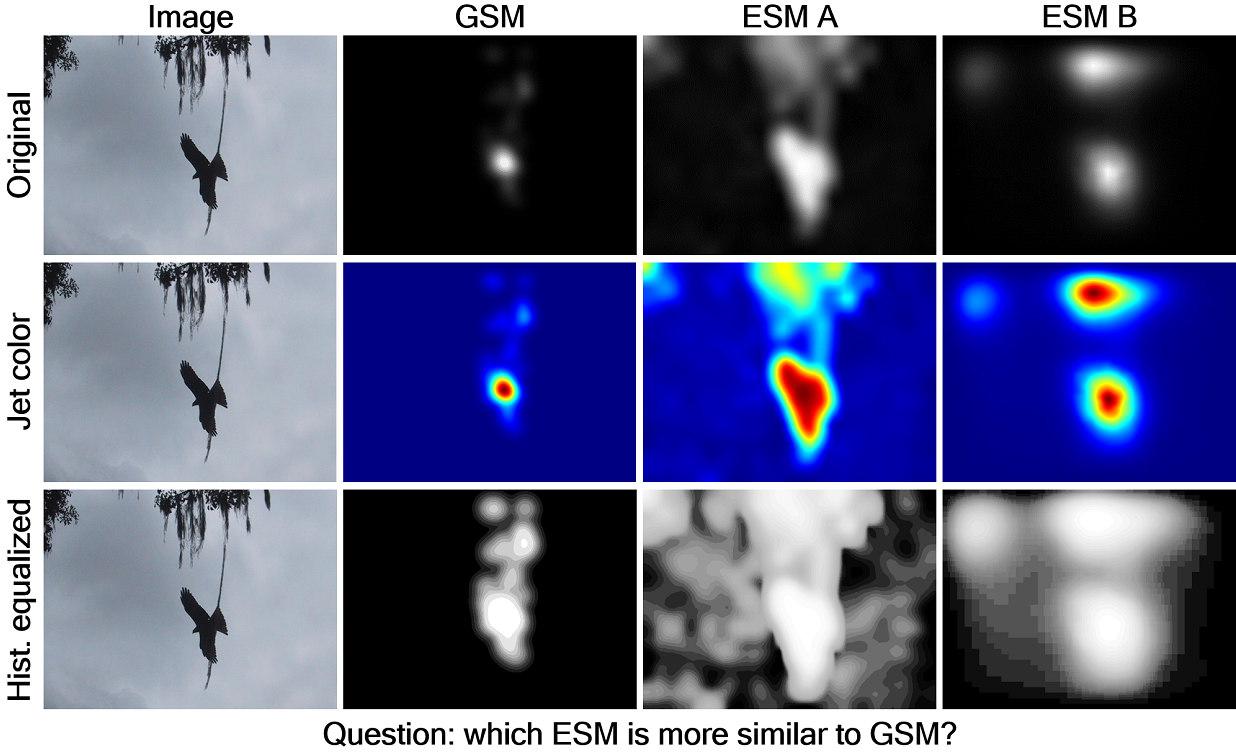}
\caption{Three different  ways of visual comparison of saliency maps. The rows show the original fixation distribution, the  jet color maps, and the histogram equalized maps, respectively. In our subjective tests, we design a set of questions each of which displays the original image, one ground-truth saliency map (GSM) and two estimated saliency maps (ESMs), where both GSM and ESMs are alternatively visualized with the histogram equalized maps as shown in the third row. Subjects are asked to determine which ESM is more similar to the displayed GSM. In this manner, the crowdsourced data collected from subjective tests contain useful cues about the human perception of saliency maps, which can be used to train a evaluation metric for measuring perceptual similarity of saliency maps .}
\label{fig:questionnaire}
\end{figure}

Toward this end, researchers have proposed various metrics. For example, the Area Under the ROC Curve, referred to as~\bl{AUC}, is a popular metric to measure the tradeoff between true and false positives at various discrimination thresholds applied to the saliency map~(\eg, \cite{harel2007graph,judd2009learning,li2010probabilistic}); the Normalized Scanpath Saliency (\bl{NSS}) aims to measure visual saliency at fixated locations so as to be invariant to linear transformations~\cite{peters2005components}; the Similarity metric (\bl{SIM}) is introduced to indicate the intersection between pairs of normalized saliency distributions~\cite{judd2012benchmark}; Kullback-Leibler Divergence (\bl{KLD}) is regarded as a measure of the information loss by evaluating saliency with a probabilistic interpretation~\cite{rajashekar2004point,tatler2005visual}; and the Earth Movers's Distance (\bl{EMD}) measures the cost transforming one distribution to another. All of these metrics evaluate saliency models quantitatively by calculating the performance scores~\cite{rubner2000earth}.

Beyond these quantitative metrics, the qualitative visual comparison from multiple models is also adopted by almost all previous works. In~\figref{fig:questionnaire}, three different visualization ways of direct visual comparison of saliency maps are illustrated. Beginning with the influential model of~\cite{itti1998model}, the original fixation distributions are displayed to qualitatively compare the performance of the corresponding saliency models~(\eg,~\cite{mit-saliency-benchmark,zhang2008sun,harel2007graph,zhang2013boolean}), but small values in the original fixation maps are not perceptible so that it can not always enter into the comparison process, such as the tree branches shown in the first row of~\figref{fig:questionnaire}. In view of this point, the jet color maps shown in the second row are utilized to make a better perceptual judgment~(\eg,~\cite{borji2013state,borji2013quantitative,russell2014model}). Moreover, Bruce \etal~\cite{bruce2015computational} propose a comparison based on histogram equalization as shown in the third row, wherein the spread of saliency values produced by each model is mapped as closely as possible into the same space. Actually, the human perception of visual similarity underlying these direct visual comparisons is very helpful to design new metrics that qualitatively assess saliency maps as humans do.

To investigate the key factors that influence the human perception of visual similarity in comparing the saliency maps, we conduct subjective tests. As shown in~\figref{fig:questionnaire}, we ask multiple subjects to determine which of the two estimated saliency maps (ESMs) is more similar to the ground-truth saliency map (GSM). How to visualize GSM and ESMs plays an important role in perceptual judgment of maps. It is difficult to make direct comparisons using the original fixation maps as shown in the first row since the maps vary greatly in their amount of salient pixels. Different from our previous work~\cite{li2015metric} which displays only ESMs and GSMs with jet colors as shown in the second row in~\figref{fig:questionnaire}, here we display histogram equalized maps in the third row so that small values also become perceptible and enter into the visual comparison process. In addition, the original image is displayed as well to facilitate the visual comparison process. We collect $134,400$ binary annotations from 16 subjects. Through the analyses of the comparison process, we find four key factors that may affect the evaluation of saliency maps. Most importantly, there indeed exists consistency among the human subjective judgments.

Based on these findings and the crowdsourced data, we propose to learn a CNN-based saliency evaluation metric using crowdsourced perceptual judgements, and such a metric is abbreviated as \bl{CPJ}. To optimize the parameters of \bl{CPJ}, we design a two-steam CNN architecture, within which each stream corresponds to the same \bl{CPJ} metric (\ie, two CNNs with cloned parameters). The whole architecture takes two ESMs (denoted as $A$ and $B$) and one GSM (denoted as $G$) as the input, while its two streams focus on predicting the relative saliency scores of $(A,G)$ and $(B,G)$, respectively. After that, the difference between the scores given by the two streams is expected to approximate the crowdsourced perceptual judgement, which is represented by a numerical score to depict the relative performance of $A$ and $B$ in approximating $G$. Finally, each stream can be viewed as a evaluation metric that assigns a numerical performance score to reveal the perceptual similarity between an ESM and the corresponding GSM, which is the same as existing classic metrics. Compared with the ten representative metrics, experimental results show that the CNN-based metric has the highest consistency with the human perception in comparing the visual similarity between ESMs and GSM, making it a good complement to existing metrics. Besides, due to the effectiveness and characteristic of the learned metric, it can be used to develop new saliency models.

The main contributions of this paper include:

1)~We collect massive crowdsourced data through subjective tests, based on which the performance of evaluation metrics in approximating human perception of saliency maps are directly quantized and compared. %These user data will be released to facilitate the assessment of existing metrics and the design of new ones.

2)~The performance of ten representative metrics are quantized for direct comparison at image and model levels. Both perspectives prove that there still exists a large gap between the inherent characteristics of existing metrics and the human perception of visual similarity.

3)~We propose a CNN-based metric that agrees better with perceptual similarity judgments of saliency maps. Experimental results show that the learned metric can be utilized for the assessment of saliency maps from new images, new datasets, new models and synthetic data. %The effectiveness of the learned metric revealed can be used to facilitate the development of new models for fixation prediction.

The rest of this paper is organized as follows: Section~\uppercase\expandafter{\romannumeral2} discusses related works. Section~\uppercase\expandafter{\romannumeral3} presents details in subjective tests and analyzes ten representative metrics. Section~\uppercase\expandafter{\romannumeral4} proposes how to learn a comprehensive evaluation metric with CNNs. Extensive experiments are conducted in Section~\uppercase\expandafter{\romannumeral5} to validate the learned metric. Finally, the paper is concluded in Section~\uppercase\expandafter{\romannumeral6}.

\section{Related Work}\label{sec:survey}

In the literature, there already exist several surveys of saliency models and  evaluation metrics (\eg, \cite{borji2013analysis,emami2013selection,riche2013metric}). Here, we first briefly introduce ten representative metrics that are widely used in existing studies. Besides, metric analyses are discussed in some references.

\subsection{Classic Evaluation Metrics}

Let $S$ be an ESM and $G$ be the corresponding GSM, some metrics select a set of positives and/or negatives from $G$, which are then used to validate the predictions in $S$. Representative metrics that adopt such an evaluation methodology include $\phi_1$ to $\phi_5$, explained below.

\myPara{Area Under the ROC Curve} (\bl{AUC}, $\phi_1$). \bl{AUC} is a classic metric widely used in many works (\eg, \cite{harel2007graph,judd2009learning,li2010probabilistic}). It selects all the fixated locations as positives and takes all the other locations as negatives. Multiple thresholds are then applied to $S$, and the numbers of true positives, true negatives, false positives and false negatives are computed at each threshold. Finally, the ROC curve can be plotted according to the true positive rate and false positive rate at each threshold. Perfect $S$ leads to an \bl{AUC} of 1, while random prediction has an \bl{AUC} of 0.5. In this study, we adopt the implementation from \cite{judd2009learning} to compute \bl{AUC}.

\myPara{Shuffled AUC} (\bl{sAUC}, $\phi_2$). Since fixated locations often distribute around image centers (\ie, the center-bias effect), \bl{AUC} favors saliency models that emphasize central regions or suppress peripheral regions. As a result, some models gain a remarkable improvement in \bl{AUC} by simply using center-biased re-weighting or border-cut. To address this problem, \bl{sAUC} selects negatives as the fixated locations shuffled from other images in the same benchmark (\eg, \cite{hou2012image,lu2013robust,zhang2013boolean}). Since both fixated and non-fixated locations are both center-biased, simple center-biased re-weighting or border-cut will not dramatically change the performance in \bl{sAUC}. In this study, we adopt the implementation from \cite{zhang2013boolean} to compute \bl{sAUC}.

\myPara{Resampled AUC} (\bl{rAUC}, $\phi_3$). One drawback of \bl{sAUC} is that label ambiguity may arise when adjacent locations in images are simultaneously selected as positives and negatives (\eg, locations from the same object). Due to the existence of such ambiguity, even the GSM $G$ cannot reach a \bl{sAUC} of 1, and such ``upper-bound'' may change on different images. To address this problem, Li~\etal~\cite{li2014visual} proposed to re-sample negatives from non-fixated locations (\ie, regions in $G$ with low responses) according to the fixation distribution over the whole dataset. In this manner, the selected positives and negatives have similar distributions, and the ambiguity can be greatly alleviated in computing \bl{rAUC}. In this study, we adopt the implementation of \bl{rAUC} from \cite{li2016measuring,li2015finding}, which selects the same amount of positives and negatives from each image.

\myPara{Precision} (\bl{PRE}, $\phi_4$). Metrics such as \bl{AUC}, \bl{sAUC} and \bl{rAUC} mainly focus on the ordering of saliency and ignore the magnitude~\cite{margolinevaluate,zhao2012learning}. To measure the saliency magnitudes at positives, \bl{PRE} was proposed in \cite{li2012removing,li2016measuring} to measure the ratio of energy assigned only to positives (\ie, fixated locations, and \bl{PRE} is denoted as Energy-on-Fixations in \cite{li2015finding}). A perfect $S$ leads to a \bl{PRE} score of 1, and a ``clean'' $S$ usually has a higher \bl{PRE} score. In this study, we select positives and negatives as those used in computing $\bl{rAUC}$, which is also the solution of \cite{li2015finding}.

\myPara{Normalized Scan-path Saliency} (\bl{NSS}, $\phi_{5}$). To avoid selecting negatives, \bl{NSS} only selects positives (\ie, fixated locations~\cite{erdem2013visual,peters2009congruence}). By normalizing $S$ to zero mean and unit standard deviation, \bl{NSS} computes the average saliency value at selected positives. Note that \bl{NSS} is a kind of Z-score without explicit upper and lower bounds. The larger \bl{NSS}, the better $S$.

Instead of explicitly selecting positives and/or negatives, some representative metrics propose to directly compare $S$ and $G$ by taking them as two probability distributions. Representative metrics that adopt such an evaluation methodology include $\phi_6$ to $\phi_{10}$, explained below.

\myPara{Similarity} (\bl{SIM}, $\phi_{6}$). As stated in \cite{hou2013visual}, \bl{SIM} can be computed by summing up the minimum value at every location of the saliency maps $S$ and $G$, while $S$ and $G$ are both normalized to sum up to one. From this definition, \bl{SIM} can be viewed as the intersection of two probability distributions, which falls in the dynamic range of $[0,1]$. Larger \bl{SIM} scores indicate better ESMs.

\myPara{Correlation Coefficients} (\bl{CC}, $\phi_{7}$). \bl{CC} describes the linear relationship between two variables \cite{borji2012boosting,lang2012saliency}. It has a dynamic range of $[-1,1]$. Larger \bl{CC} indicates a higher similarity between $S$ and $G$.

\myPara{Information Gain} (\bl{IG}, $\phi_{8}$). \bl{IG}, as an information theoretic metric, is defined as the entropic difference between the prior and the posterior distribution~\cite{Kummerer2014a,Kuemmerer2015a}. \bl{IG} is like \bl{KLD} but baseline-adjusted. In~\cite{BylinskiiJOTD16}, \bl{IG} over a center prior baseline provides a more intuitive way to interpret model performance relative to center bias.

\myPara{Kullback-Leibler Divergence} (\bl{KLD}, $\phi_{9}$). \bl{KLD} is an entropy-based metric that directly compares two probability distributions. In this study, we combine the \bl{KLD} metrics in \cite{borji2013state} and \cite{riche2013metric} to compute a symmetric \bl{KLD} according to the saliency distributions over $S$ and $G$. In this case, smaller \bl{KLD} indicates a better performance.

\myPara{Earth Mover's Distance} (\bl{EMD}, $\phi_{10}$). The \bl{EMD} metric measures the minimal cost to transform one distribution to the another~\cite{erdem2013visual,zhao2012learning}. Compared with $\phi_1-\phi_9$, the computation of \bl{EMD} is often very slow since it requires complex optimization processes. Smaller \bl{EMD} indicates a better performance.

Most existing works on fixation prediction adopted one or several metrics among $\phi_1-\phi_{10}$ for performance evaluation. In these works, we notice that the resolutions of $S$ and $G$, as well as the interpolation methods to down-sample or up-sample $S$ and $G$ to the same resolutions, may change the scores of some metrics. Therefore, when using $\phi_1-\phi_{10}$, we up-sample or down-sample $S$ to the same size of $G$ by bilinear interpolation. After that, we normalize $S$ and $G$ to have the minimum value 0 and the maximum value 1.

\subsection{Metric Analysis}

Given these representative metrics, there also exist some works in metric analysis. Wilming~\etal~\cite{wilming2011measures} explored how models of fixation selection can be evaluated. Through deriving a set of high-level desirable properties for metrics through theoretical considerations, they analyzed eight common measures with these requirements and concluded that no single measure can capture them all. Then both \bl{AUC} and \bl{KLD} were recommended to facilitate comparison of different models and to provide the most complete picture of model capabilities. Regardless of the measure, they argued that model evaluation was also influenced by inherent properties of eye-tracking data.

Judd~\etal~\cite{judd2012benchmark} provided an extensive review of the important computational models of saliency and quantitatively compared several saliency models against each other. They proposed a benchmark data set, containing 300 natural images with eye tracking data from 39 observers to compare the performance of available models. For measuring how well a model predicted where people look in images, three different metrics including \bl{AUC}, \bl{SIM} and \bl{EMD} were utilized to conduct a more comprehensive evaluation. Besides, they showed that optimizing the model blurriness and bias towards the center to models improves performance of many models.

Emami~\etal~\cite{emami2013selection} proposed to identify the best metric in terms of human consistency. By introducing a set of experiments to judge the biological plausibility of visual saliency models, a procedure was proposed to evaluate nine metrics for comparing saliency maps using a database of human fixations on approximately 1000 images. This procedure was then employed to identify the best saliency comparison metric as the one which best discriminates between a human saliency map and a random saliency map, as compared to the ground truth map.

Riche~\etal~\cite{riche2013metric} investigated the characteristics of existing metrics. To show which metrics are closest to each other and see which metric should be used to do an efficient benchmark, Kendall concordance coefficient is used to compare the relative rank of the saliency models according to the different metrics. Based on the nonlinear correlation coefficient, it is shown that some of the metrics are strongly correlated leading to a redundancy in the performance metrics reported in the available benchmarks. As a recommendation, \bl{KLD} and \bl{sAUC} are most different from the other metrics, including \bl{AUC}, \bl{CC}, \bl{NSS}, and \bl{SIM}, which formed a new measure.

Bruce~\etal~\cite{bruce2015computational} argued that there existed room for further efforts addressing some very specific challenges for assessing models of visual saliency. Rather than considering fixation data, annotated image regions and stimulus patterns inspired by psychophysics to assess the performance benchmark of saliency models under varying conditions, they aimed to present a high level perspective in computational modeling of visual saliency with an emphasis on human visual behavior and neural computations. They analyzed the shortcomings and challenges in fixation-based benchmarking motivated by the spatial bias, scale and border effect. Besides, they further discussed the biological plausibility of models with respect to behavioral findings.

K{\"u}mmerer~\etal~\cite{kummerer2015information} argued that a probabilistic definition is most intuitive for saliency models and explored the underlying reason why existing model comparison metrics give inconsistent results. They
offered and information theoretic analysis of saliency evaluation by framing fixation predication models probabilistically and introduced the notion of information gain. Toward this end, they proposed to compare models by jointly optimizing factors such as scale, center bias and spatial blurring so as to obtain consistent model ranking across metrics. Besides, they provided a method to show where and how saliency models fail. % to capture information in the fixation on the pixel level.

%On the issues about the saliency models receive different ranks according to different evaluation metrics,
Bylinskii~\etal~\cite{BylinskiiJOTD16} aimed to understand the essential reason why saliency models receive different ranks according to different evaluation metrics. Rather than providing tables of performance values and literature reviews of metrics, they offered a more comprehensive explanation of 8 common evaluation metrics and present visualization of the metric computations. By experimenting on synthetic and natural data, they revealed the particular strengths and weaknesses of each metric. Besides, they analyzed these metrics under various conditions to study their behavior, including the treatment of false positives and false negatives, systematic viewing biases, and relationship between metrics by measuring how related the rankings of the saliency models are across metrics. Building on the results of their analyses, they offered guidelines for designing saliency benchmarks and choosing appropriate metrics.

Although these works provide some insights into the advantages and disadvantages of existing representative quantitative metrics, it is still important to obtain a quantitative metric that performs more consistently with humans in assessing visual saliency models. In our previous work~\cite{li2015metric}, we conducted extensive subjective tests and proposed a data-driven metric using convolutional neural networks. Compared with existing metrics, the data-driven metric performs most consistently with the humans in evaluating saliency maps as well as saliency models. However, the jet colormap chosen for subjective tests may fall very far from having distances in colorspace that are a good match for perceptual distances. Besides, the evaluation depends on performing pairwise comparisons, which adds additional complications to assess the saliency model benchmarking.

Toward this end, we consider more reasonable perceptual factors and propose to directly quantize the performance of metrics by using the crowdsourced data collected from extensive subjective tests. Based on the crowdsourced perceptual judgements, a saliency evaluation metric is then learned by using the CNNs, which can qualitatively measure human perceptual similarity of saliency maps.

\section{Subjective Tests for Metric Analysis}\label{sec:expStudy}

In this section, we conduct subjective tests to study how saliency map are compared by the humans. Based on the crowdsourced data collected in these tests, we carry out extensive image-level and model-level analyses to quantize and compare the performance of existing metrics in terms of measuring the human perception of visual similarity.

\subsection{Subjective Tests}

\begin{figure}[t]
\centering
\includegraphics[width=0.50\textwidth]{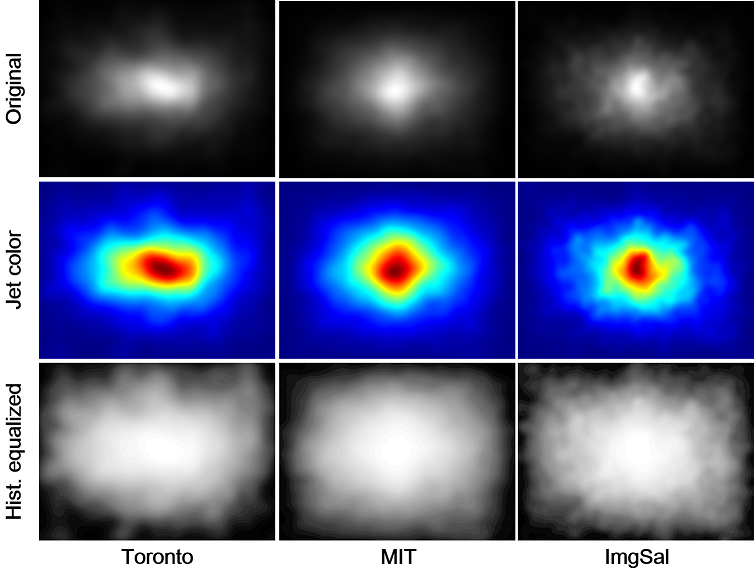}
\caption{Average fixation density maps of three datasets. The first row shows the original fixation distribution and the second row shows the jet color map, while the third row displays the histogram equalized maps to get a better perceptual comparisons \cite{judd2012benchmark}.}
\label{fig:AVGMaps}
\end{figure}

In subjective tests, we select 400 images from three datasets, including 120 images from \bl{Toronto}~\cite{bruce2005saliency}, 180 images from \bl{MIT}~\cite{judd2009learning} and 100 images from \bl{ImgSal}~\cite{li2013visual}. Human fixations on these image are collected by different eye-tracking configurations, leading to lower dataset bias. For each image, we generate seven ESMs with seven saliency models, including $\mc{M}_0$ (AVG), $\mc{M}_1$ (IT~\cite{itti1998model}), $\mc{M}_2$ (GB~\cite{harel2007graph}), $\mc{M}_3$ (CA~\cite{goferman2012context}), $\mc{M}_4$ (BMS~\cite{zhang2013boolean}), $\mc{M}_5$ (HFT~\cite{li2013hypercomplex}) and $\mc{M}_6$ (SP~\cite{li2014visual}). Note that AVG simply outputs the average fixation density map from \bl{Toronto}, \bl{MIT} and \bl{ImgSal} (see \figref{fig:AVGMaps}). For each image, the 7 ESMs form $C_7^2=21$ ESM pairs. Based on the ESM pairs, we carry out subjective tests with $400\times{}21=8,400$ questions.

Typically, the choice of colormap can impact significantly the human perception of visual similarity and can also play a role in shaping how saliency maps are judged, and what information is discernible. In our previous work~\cite{li2015metric}, each question only consisted of two ESMs and one GSM displayed in jet color. Subject needs to determine which ESM is more similar to GSM, without knowing the models that generate the ESMs. In total, 22 subjects (17 males and 5 females, aged from 22 to 29) participated in the tests. Note that each question is presented to exactly 8 subjects, and all subjects know the meaning of colors in ESMs and GSMs (i.e., which colors correspond to the most salient locations and which colors indicate background regions). In the subjective tests, there is no time limitation for a subject in answering a question. Finally, we obtain $8400\times{}8=67,200$ answers (i.e., binary annotations).

Based on the annotations, the learned metric obtained an impressive performance. However, such an experimental setting has three deficiencies. First, the distance in the jet colorspace may fall a little far from perceptual distances. Second, the original image is not displayed to facilitate the visual comparison, which is actually a necessary supplement when both ESMs are not very visually similar to the GSM. Third, it is hard to judge whether the number of subjects presented to each question is sufficient to make proper annotations.

To address these, we adopt a new setting in the subjective tests by displaying the original image in each question and visualizing histogram-equalized gray saliency maps to make better perceptual comparisons, as suggested in \cite{bruce2015computational}. Finally, each of the 8,400 questions conducted in the current subjective tests consists of one color image as well as one GSM and two ESMs visualized as histogram-equalized gray saliency maps (see the third row of Fig.~\ref{fig:questionnaire}).

In total, 16 subjects (15 males and 1 female, aged from 21 to 24) participated in the tests. All subjects had normal or corrected to normal vision. In the tests, each subject was requested to answer all questions with at least 5 seconds per question. Finally, we obtained $8,400\times{}16=134,400$ answers (binary annotations) under this setting. For the sake of simplification, we represent the crowdsourced data as
\begin{equation}
\left\{\left(A_{k},B_{k},G_k\right),l_k, c_k, r_k|k\in\mb{I}\right\},
\end{equation}
where $\mb{I}=\{1,\ldots,8400\}$ is the set of question indices. $A_{k}$ and $B_{k}$ are two ESMs being compared with the GSM $G_k$ in the $k$th question. $l_k\in[0,1]$ is computed as the percentage of subjects that choose $A_k$ in answering the $k$th question. We can see that one ESM clearly outperforms the other one when $l_k$ equals to 0 or 1, while $l_k=0.5$ indicates two ESMs with similar performance. For the sake of simplification, we present a variable $c_k\in[0,1]$ which is computed as $2\times|l_k-0.5|$ to quantize the confidence that one ESM clearly outperforms the other one. Besides, relative saliency score $r_k\in[-1,1]$ is the likelihood that $A_k$ outperforms $B_k$, which is computed as the percentage of difference between subjects that choose $A_k$ and $B_k$ (\ie, $r_k=2\cdot{}l_k-1$).

After the tests, subjects are also requested to explain the key factors they adopted in making decisions. By investigating their explanations, we find the following key factors that may affect the evaluation of saliency maps.

\myPara{1)~The most salient and non-salient locations}. In most cases, both ESMs can unveil visual saliency to some extent, and the most salient and non-salient regions play a critical role in determining which ESM performs better. In particular, the overlapping ratio of the most salient and non-salient regions between ESM and GSM is the most important factor in assessing saliency maps.

\myPara{2)~Energy distribution}. The compactness of salient locations is an important factor for assessing saliency maps. ESMs that only pop-out object borders are often considered to be poor. Moreover, the background cleanness also influences the evaluation since fuzzy ESMs usually rank lower in the pair-wise subjective comparisons.

\myPara{3)~Number and shapes of salient regions}. A perfect ESM should contain exactly the same number of salient regions as in the corresponding GSM. Moreover, salient regions with simple and regular shapes are preferred by most subjects.

\myPara{4)~Salient targets in the original image}. When it is difficult to judge which ESM performs better in approximating the GSM, subjects may refer to the original image and check what targets actually pop-out in both ESMs.

\subsection{Statistics of User Data}

\begin{figure}[t]
\centering
\includegraphics[width=0.48\textwidth]{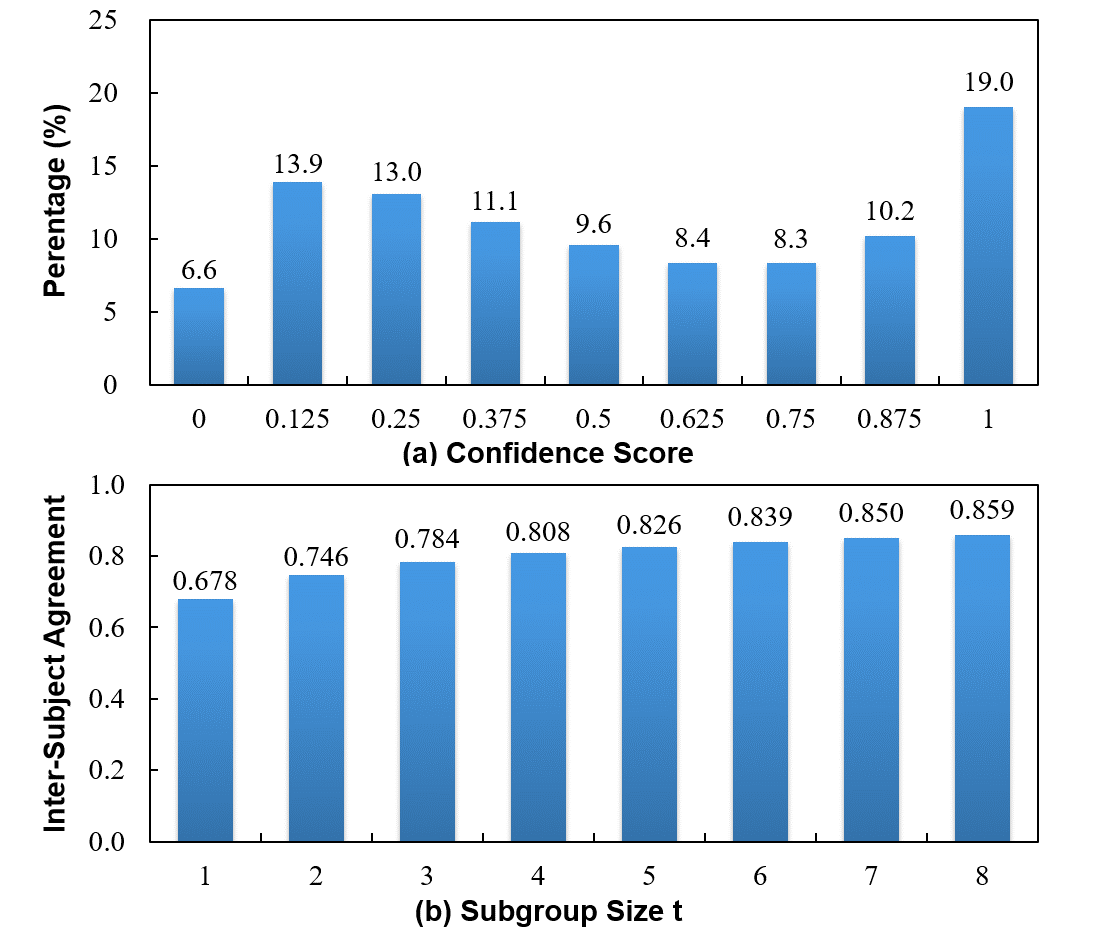}
\caption{Statistics of user data. (a) Distribution of confidence scores over 8,400 questions. (b) Inter-subject agreement between different subgroups.}
\label{fig:dataStat}
\end{figure}

\begin{figure}[t]
\centering
\includegraphics[width=0.48\textwidth]{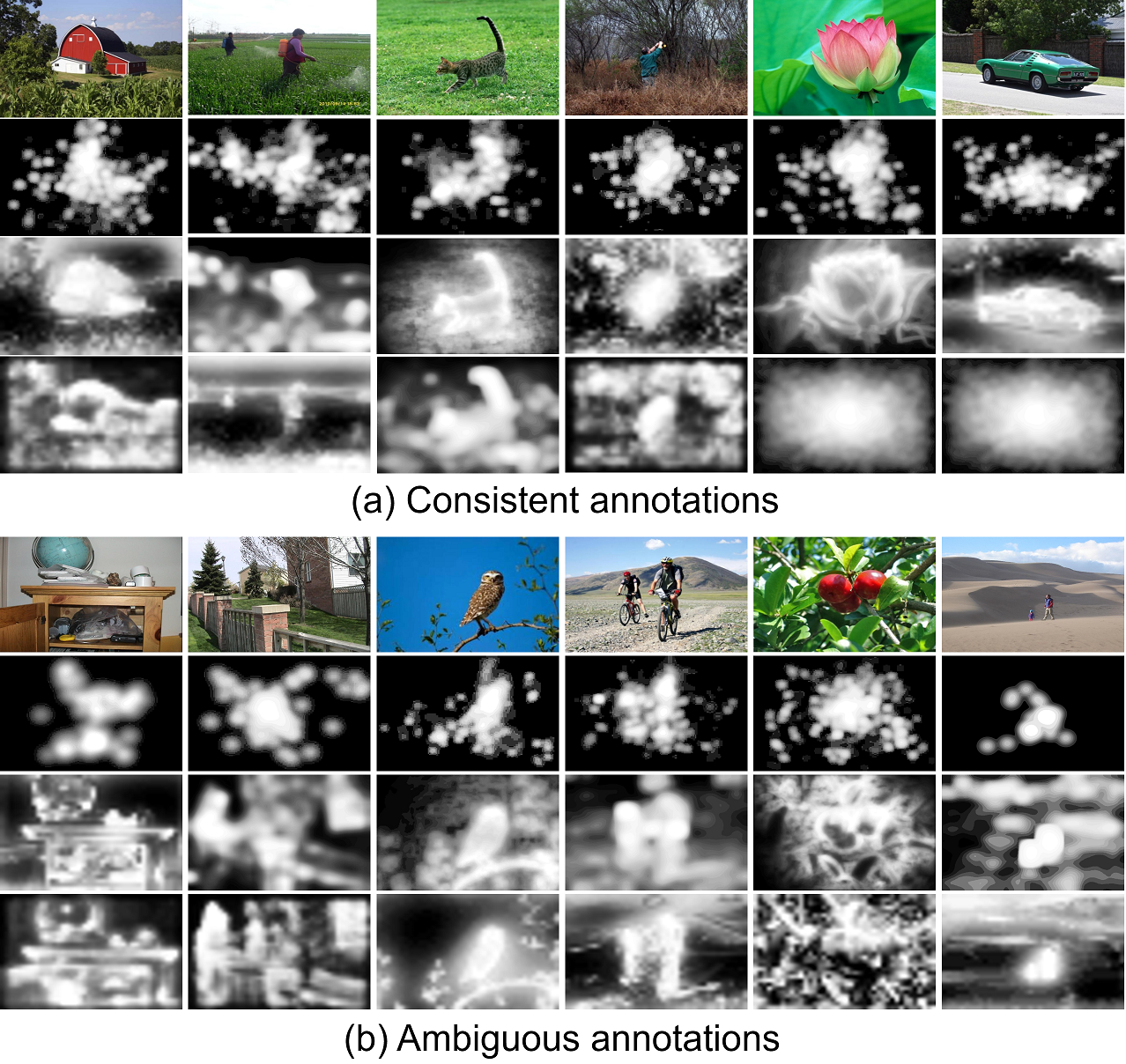}
\caption{Representative examples of consistent and ambiguous annotations ($1$st row: Original image; $2$nd row: GSM; $3$rd and $4$th rows: ESMs). In (a), ESMs at the $3$rd row outperform those at the $4$th row. In most of the cases, ESMs from $\mc{M}_1-\mc{M}_6$ outperform
ESMs from AVG (the last two columns of (a)).}
\label{fig:specialCases}
\end{figure}

Given the crowdsourced data collected from subjective tests, an important concern is whether the annotations from various subjects are consistent. To answer this question, we defined two types of annotations in our previous work~\cite{li2015metric} and found that there indeed existed consistency in most cases. In this work, we show the distribution of confidence scores $\left\{c_{k}|k\in\mb{I}\right\}$ in Fig.~\ref{fig:dataStat} (a). From Fig.~\ref{fig:dataStat} (a), we find that the majority of subjects act highly consistently in answering about 79.5\% of questions with confidence scores no less than 0.25 (\ie, at least 10 out of the 16 subjects select the same ESM in answering a question), indicating that there do exist some common clues among different subjects in assessing the quality of saliency maps. We define this type of annotations as consistent annotations. As shown in \figref{fig:specialCases} (a), such annotations often occur when one ESM performs significantly better than the other one. Meanwhile, we also notice that in 20.5\% of questions the annotations are quite ambiguous with confidence scores below 0.25, which are defined as ambiguous annotations, making it difficult to distinguish which ESM performs better. As shown in \figref{fig:specialCases} (b), both ESMs in most of these questions perform unsatisfactory and it is difficult to determine which ESM is better.

Beyond the consistency, another concern is that whether 16 subjects presented to each question are enough or not. To answer this, we compute the inter-subject agreement between different groups of subjects. We divide the 16 subjects into a set of subgroups $\mb{U}_t$ with group size $t\in\{1,\ldots,8\}$, and the inter-subject agreement at group size $t$, denoted as $\alpha_t$, can be computed as
\begin{equation}
\alpha_t=1-\frac{\sum_{u_1,u_2\in{}\mb{U}_t}\xi(u_1\cap{}u_2=\varnothing)\cdot\sum_{k\in\mb{I}}|l_k^{u_1}-l_k^{u_2}|}{\sum_{u_1,u_2\in{}\mb{U}_t}\xi(u_1\cap{}u_2=\varnothing)\cdot|\mb{I}|}
\end{equation}
where $u_1$ and $u_2$ are two subgroups formed by different subjects. $l_k^{u_1}$ and $l_k^{u_2}$ are the likelihood that the ESM $A_k$ outperforms $B_k$ in the subgroups $u_1$ and $u_2$, respectively. By enumerating the pair-wise combination all subgroups, we show the inter-subject agreement in Fig.~\ref{fig:dataStat} (b). We can see that when the size of subgroup grows, the decision made by different subgroups gradually becomes more consistent. In particular, the growth in the inter-subject agreement stays almost stable (\ie, from 0.850 to 0.859) even when the group size varies from $t=7$ to $t=8$, implying that 16 subjects are sufficient to provide stable visual comparison results in the tests we conducted.

\begin{figure*}[t]
\centering
\includegraphics[width=0.80\textwidth]{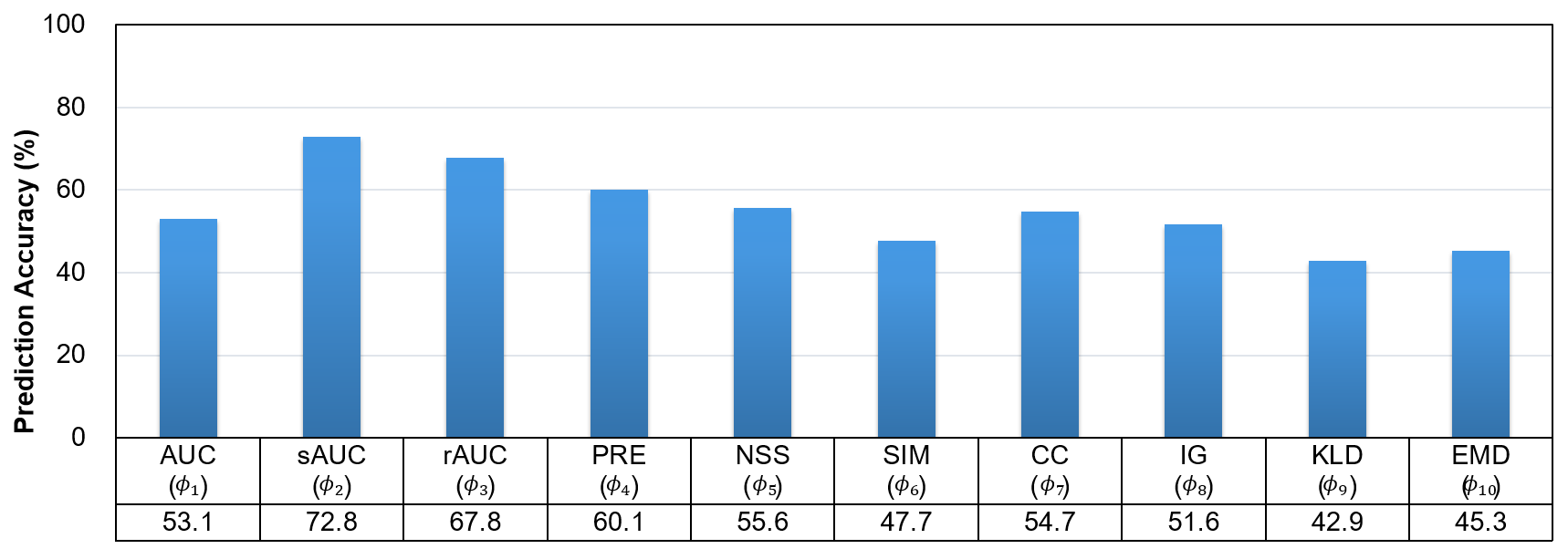}
\caption{Prediction accuracies of ten representative evaluation metrics, which indicate the consistency degree in predicting the ordering of all ESMs in terms of measuring the human perception of visual similarity.}
\label{fig:metricAccuracy}
\end{figure*}

 \subsection{Analysis of Ten Representative Metrics}

Given the crowdsourced data, we can quantize the performance of $\phi_1-\phi_{10}$ so as to directly compare them in terms of measuring the human perception of visual similarity. The comparisons are conducted from two perspectives, including image-level and model-level. In image-level comparison, we aim to see if existing metrics can correctly predict which ESM acts better similar to humans. Given a metric $\phi_i$, its accuracy in predicting the ordering of ESMs can be computed as:
\begin{equation}\label{eq:predictionAccuracy}
  P_i=\frac{1}{\sum_{k\in\mb{I}}c_k}\cdot\left\{
\begin{array}{c@{\;,\;}l}
\sum_{k\in\mb{I}}[cs_k>0]_\text{I}\cdot{c_k}& i=1,...,8\\
\sum_{k\in\mb{I}}[cs_k<0]_\text{I}\cdot{c_k}& i=9,10
\end{array}
\right.
\end{equation}
where $cs_k=(\phi_i(A_{k})>\phi_i(B_{k}))\cdot(l_k-0.5)$, and $[\bl{e}]_\text{I}=1$ if the event \bl{e} holds, otherwise $[\bl{e}]_\text{I}=0$.  Note that in \eqref{eq:predictionAccuracy} we omit $G$ in $\phi_i(S,G)$.

The accuracies of ten heuristic metrics are shown in Fig.~\ref{fig:metricAccuracy}. We find that the top two metrics that perform most consistently with human perception are $\phi_{2}$ (\bl{sAUC}) and $\phi_3$ (\bl{rAUC}) and the lowest prediction accuracy is only 42.9\% from $\phi_9$ (\bl{KLD}). However, the best metric, \bl{sAUC}, only reaches an accuracy of 72.8\% in comparing all the ESM pairs, while random prediction achieves an accuracy of 50\% in addressing such binary classification problems. Actually, there still exists a huge gap between these existing metrics and the human perception of visual similarity. Note that \textit{the low accuracy does not mean that the metric is not suitable for assessing saliency models}. Instead, such a low-accuracy metric works complementary to the direct visual comparisons of histogram equalized saliency maps.

\begin{figure}[t]
\centering
\includegraphics[width=1.00\columnwidth]{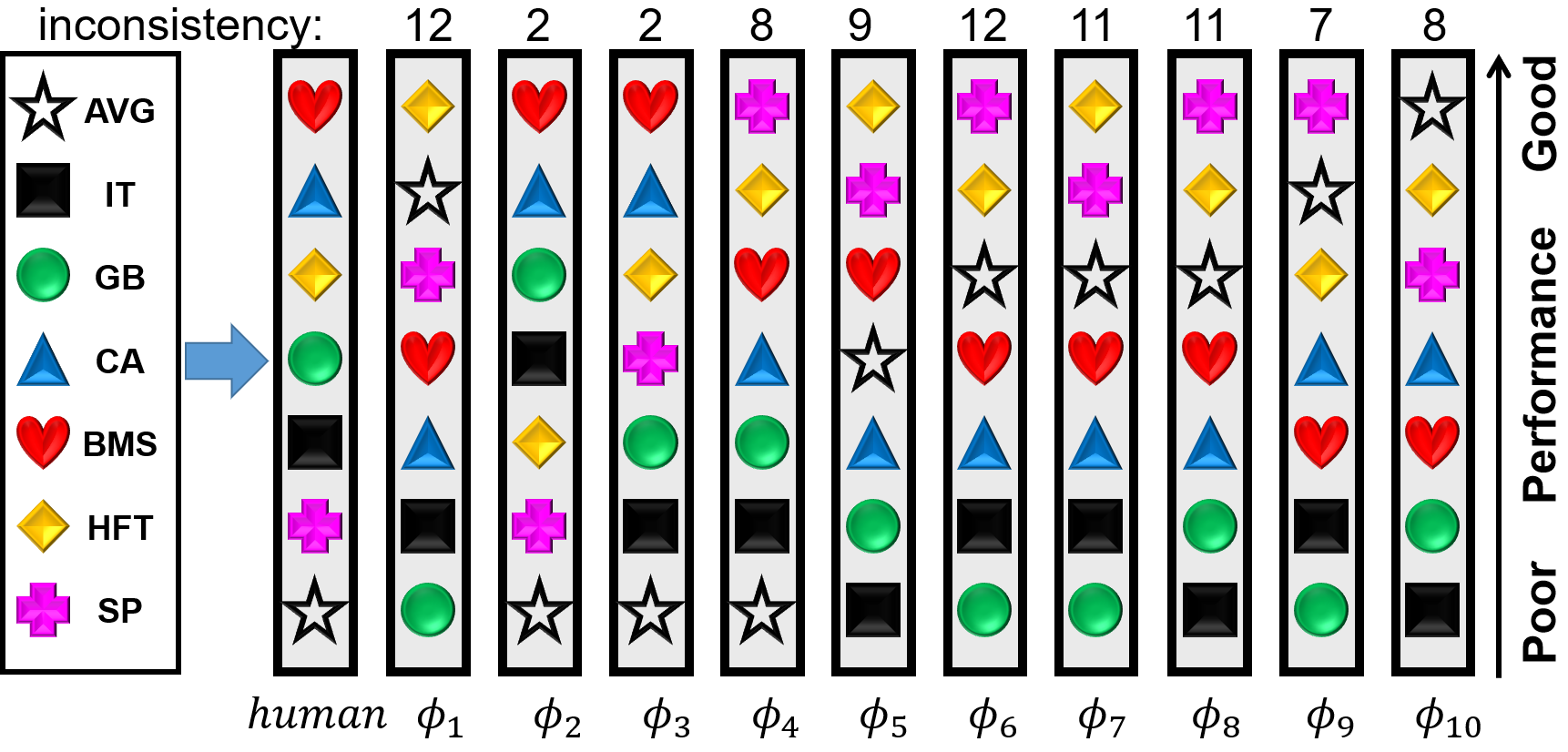}
\caption{Ranking lists of seven models generated by 10 representative evaluation metrics. The number above each bar indicates how many pairs of models are inconsistently ranked. }
\label{fig:modelRanking}
\end{figure}

In addition to the image-level comparison, we also compare these ten metrics at the model-level. That is, we generate a ranking list of the seven models with each metric. The model rankings generated by various metrics, as well as the numbers of inconsistently predicted model pairs, can be found in \figref{fig:modelRanking}. We find that $\phi_{2}$ (\bl{sAUC}) and $\phi_{3}$ (\bl{rAUC}) still perform the best. These results are almost consistent with those in the image-level comparison. In particular, these representative evaluation metrics have their respective different ranking lists. This implies that different metrics capture different characteristics of visual saliency in assessing saliency models.

From above, we find that most of existing metrics demonstrate a large inconsistency with humans. Therefore, it is necessary to develop a new metric that better captures human perception of saliency.

% 1 we aim to see how well existing metrics match with the human-being. 2 Toward this end, based on ESM pairs with high confidence scores. 3 $|\mb{I}|=8,400$ is the number of all ESM pairs.

\section{Learning a Saliency Evaluation Metric Using Crowdsourced Perceptual Judegements}\label{sec:approach}
According to the quantized performance, we need a good complement to existing metrics from the perspective of the human perception of visual similarity. Toward this end, we propose to learn a saliency evaluation metric $\phi_{CPJ}(S,G)$ from the ESM pairs judged by 16 subjects under a new setting. Different from our previous work~\cite{li2015metric} that focuses on the ordering of the ESM pair, we treat the learned metric as a regressor rather than a binary classifier. That is, the learned metric assigns each saliency map a numerical performance score similar to existing classic metrics.

%Rather than directly training the metric in an end-to-end way, we construct the Relative-Saliency network so as to optimize its parameters by using the relative-saliency score of the ESM pairs.

\begin{figure*}[t]
\centering
\includegraphics[width=1.0\textwidth]{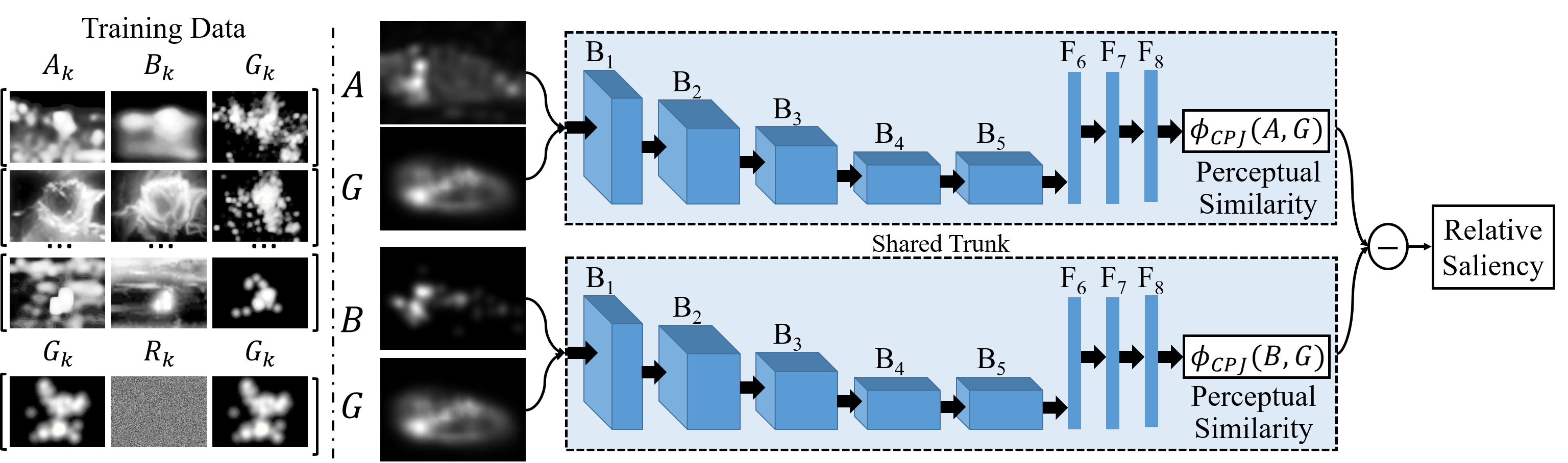}
\caption{Architecture of the Saliency Evaluation Metric based on Convolutional Neural Network for Measuring Perceptual Similarity (\bl{CPJ}). Here, The network consists of two-steam CNNs that share the same parameters and takes one ESM ($A$ or $B$) and one GSM ($G$) as the input. Each of the trunks assigns each saliency map a numerical performance score of perceptual similarity. Two separate scores after the sigmoid layers are computed by the eltwise layer with the \bl{SUM} operation, where the coeffs are set to be 1 and -1, respectively.}
\label{fig:structureCNN}
\end{figure*}

The architecture used in optimizing the parameters of \bl{CPJ} is shown in Fig.~\ref{fig:structureCNN}. It starts with two streams with cloned parameters and ends with the relative saliency performance score. Each stream is initialized from the VGG16 network~\cite{simonyan2014very}, which contains five blocks and three full connected layers. In particular, blocks $B_{1}$ and $B_{2}$ contain two convolutional layers with $3\times{}3$ kernels. Subsequently, each of the blocks $B_{3}$, $B_{4}$ and $B_{5}$ consists of three convolutional layers with $3\times{}3$ kernels. Note that we use rectified linear unit (ReLU) activation functions \cite{Hinton:2010:ICML} in the final convolutional layers of each block. Then, three fully connected layers $F_{6}$, $F_{7}$ and $F_{8}$ are followed. It is worth noting that we replace the last softmax layer with the sigmoid activation functions. Besides, different from the original VGG16 architecture, an ESM and the corresponding GSM are combined as the input for each stream. Note that both ESMs and GSM are resized to the same resolution of $128\times128$ through bilinear interpolation and are then normalized to the dynamic range of $[0,1]$. In this manner, the revised VGG16 architecture in each stream takes a $128\times{}128$ two-channel image as the input and outputs a performance score of perceptual similarity. Then, two separate perceptual similarity scores from the two streams are computed (\ie, $\phi_{CPJ}(A,G)$ and $\phi_{CPJ}(B,G)$), whose difference is then computed to approximate the relative saliency score (\ie, $r_k\in{}[-1,1], \forall k$) that indicates the likelihood that $A$ outperforms $B$ in approximating $G$. Here, we regard either of the two streams as our final metric $\phi_{CPJ}(S,G)$ since they are the same.

 To optimize the parameters in \bl{CPJ}, we adopt the crowdsourced data collected from all questions (\ie, $\left\{\left(A_{k},B_{k},G_k\right)|k\in\mb{I}\right\}$, with relative saliency score $r_k$). Unlike our previous work~\cite{li2015metric} that adopts the crowdsourced data only with consistent annotations, we utilize all crowdscourding data that takes into account all possibilities of the performance scores so as to make the learned metric more precise. In our experiments, the crowdsourced data $(A_{k},B_{k},G_k)$ are split into $(A_{k},G_k)$ and $(B_{k},G_k)$ and are fed into the two streams, respectively. Moreover, we expand the training data by swapping $A_{k}$ and $B_{k}$ (\ie, $\left\{\left(B_{k},A_{k},G_k\right)|k\in\mb{I}\right\}$, with relative saliency score $-r_k$). The mean square loss is utilized to optimize the final relative saliency scores.

\begin{table*}[t]
\caption{Accuracies (\%) of ten representative metrics and \bl{CPJ} on 50 images from \bl{Toronto} and \bl{MIT}.}
\centering {
\setlength{\tabcolsep}{3pt}
\begin{tabular}{cccccccccccc}

\toprule	
Metrics      & \bl{AUC} ($\phi_{1}$) & \bl{sAUC} ($\phi_{2}$) & \bl{rAUC} ($\phi_{3}$) & \bl{PRE} ($\phi_{4}$) & \bl{NSS} ($\phi_{5}$) & \bl{SIM} ($\phi_{6}$) & \bl{CC} ($\phi_{7}$) & \bl{IG} ($\phi_{8}$) & \bl{KLD} ($\phi_{9}$) & \bl{EMD} ($\phi_{10}$) &  \bl{CPJ}  \\
\midrule
Accuracy     & 56.5 & 71.6 & 67.3 & 63.9 & 57.9 & 50.9 & 55.9 & 52.8 & 48.1 & 47.9 & \bl{88.8} \\
\bottomrule
\end{tabular}
\begin{tablenotes}
\footnotesize
\item[1] $^*$ \bl{CPJ} is trained on the other 250 images from \bl{Toronto} and \bl{MIT} with crowdsourced perceptual judgements of 5,500 questions.
\end{tablenotes}
}\label{tab:onNewImage}
\end{table*}

Further, to set the upper and lower bound of the learned metric, we add an extra pair for each image, that is, $(G_{k},R_{k},G_k)$, with the setting of relative saliency score to 1. Specifically, $R_k$ is a synthetic saliency map with random pixel values uniformly sampled between $[0, 255]$, as shown in the training data of Fig.~\ref{fig:structureCNN}. Here, during the training phase we set additional loss functions for pairs $(G_{k},G_{k})$ and $(R_{k},G_{k})$, that is, $\frac{(1-\phi_{CPJ}(G_{k},G_{k}))^2}{2}$ and $\frac{(\phi_{CPJ}(R_{k},G_{k}))^2}{2}$, respectively. Then, for these pairs three loss values are summed directly. In this way, the performance score of pairs $(G_{k},G_{k})$ can reach to 1 and $(R_{k},G_{k})$ to 0.

All models are trained and tested with Caffe~\cite{jia2014caffe} on a single NVIDIA TITAN XP. In training \bl{CPJ}, we optimize the parameters for up to 20,000 iterations. On average, it takes about 83.2s per 100 iterations. We use stochastic gradient descent with a mini-batch size of 32 images and initial learning rate of 0.001, multiplying the learning rate by 0.1 when the error plateaus. Moreover, a momentum of 0.9 and a weight decay of 0.0005 are used. The testing speed of \bl{CPJ} is much faster since it only involves convolution, pooling and connection operations. On the same GPU platform, \bl{CPJ} takes about $8.5\times{}10^{-3}s$ to compare an ESM with the corresponding GSM (preloaded into memory).

\section{Experiments}\label{sec:experiment}

In this section, we conduct several experiments to validate the effectiveness of the learned metric on new images, new datasets, new models and synthetic data.

\subsection{Validation of the Learned Metric}

To validate the effectiveness of \bl{CPJ}, we need to test whether it can be generalized for the comparison of ESMs from new images, new datasets or new models. In addition, the rationality on synthetic data should be tested as well. Notably, in the testing phase there is an additional $(G_{k},R_{k},G_k)$ pair except for 21 pairs of 7 methods for each image. That is, there are 22 testing pairs per image. Toward this end, we design the following experiments:

\myPara{Performance on new images}. In the first experiment, we train \bl{CPJ} with the user data obtained on 250 images randomly selected from \bl{Toronto} and \bl{MIT} (\ie, $250\times22=5500$ training instances). The learned metric is then tested on all user data obtained on the remaining 50 images from \bl{Toronto} and \bl{MIT} (\ie, $50\times{}22=1100$ testing instances). Note that, the sequence of images selected for training and testing is the same as the setting in our previous work~\cite{li2015metric}. The main objective is to validate the effectiveness of the learned metric on new images whose GSMs are obtained under the same configurations in eye-tracking experiments. The performance of \bl{CPJ} and the representative metrics $\phi_{1}-\phi_{10}$ are shown in Table~\ref{tab:onNewImage}.
% Besides, the additional pairs $(G_{k},R_{k},G_k)$ are not included in the testing instances.

From Table~\ref{tab:onNewImage}, we can see that \bl{CPJ} reaches an accuracy of 88.8\%, while \bl{sAUC} and \bl{rAUC} perform the best among the metrics $\phi_{1}-\phi_{10}$ with the accuracy scores of 71.6\% and 67.3\%, respectively. This result proves that \bl{CPJ} can be generalized to the model comparisons on new images when human fixations are recorded using the same eye-tracking configurations (\eg, the rest 823 images from \bl{MIT} that are not used in the subjective tests).

Beyond the overall performance, Table~\ref{tab:onResolution} shows the accuracies of \bl{CPJ} by using different resolutions of ESMs and GSMs. We can see that the best performance is reached at the resolution $128\times128$. This can be explained by the fact that when the resolutions reduce to $64\times{}64$ and $32\times32$, many important details in ESMs and GSMs are missing, while such details may facilitate the measurement of similarity between ESMs and GSMs. On the contrary, when the resolution increases to $256\times256$, severe over-fitting risk may arise as the training data obtained from subjective tests may be somehow insufficient to train the rapidly growing parameters. Therefore, we use the resolution of $128\times128$ in training \bl{CPJ}.

\begin{table}[t]
\caption{Accuracies (\%) of \bl{CPJ} at different resolutions.}
\centering {
\begin{tabular}{ccccc}
\toprule	
Resolution      & $256\times256$ &  $128\times128$ & $64\times64$ & $32\times32$      \\
\midrule
Accuracy        & 87.8 & \bl{88.8} & 87.7 & 84.6 \\
\bottomrule
\end{tabular}
\begin{tablenotes}
\footnotesize
\item[1] $^*$ \bl{CPJ} is trained on 250 images from \bl{Toronto} and \bl{MIT} and tested on the remaining 50 images from \bl{Toronto} and \bl{MIT}. The structure of \bl{CPJ} is automatically fine-tuned for each resolution.
\end{tablenotes}
}\label{tab:onResolution}
\end{table}

\begin{figure}[t]
\centering
\includegraphics[width=1.00\columnwidth]{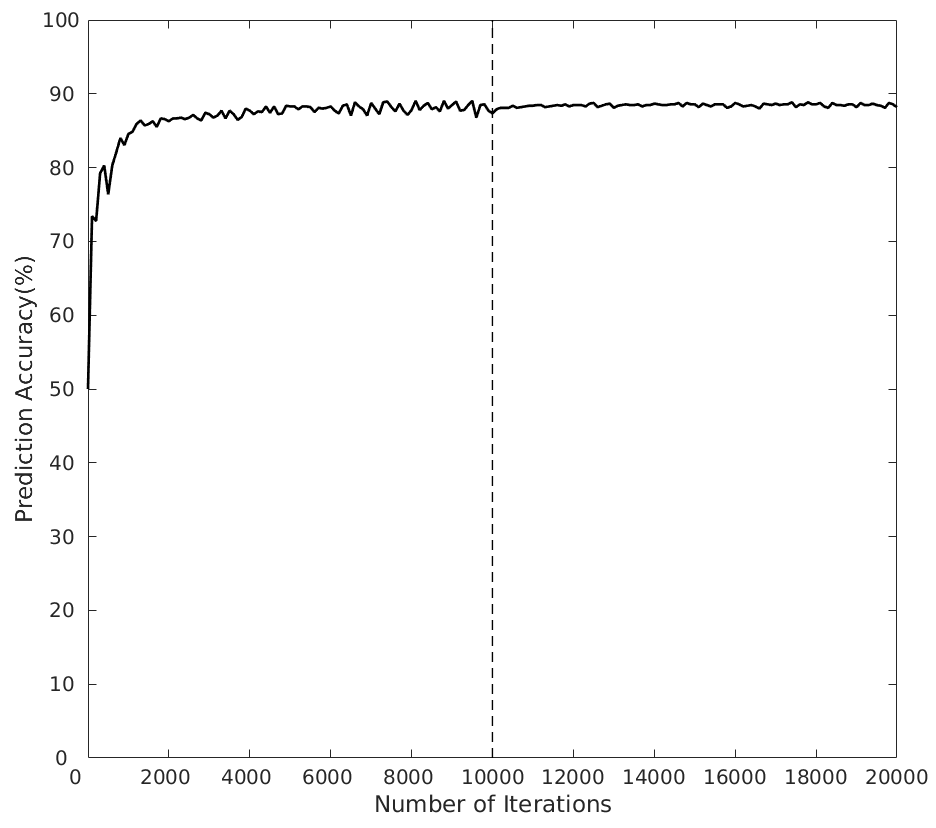}
\caption{The accuracies of \bl{CPJ} trained with different numbers of feed-forward and back-propagation iterations.}
\label{fig:iterationPerformance}
\end{figure}

Moreover, Figure~\ref{fig:iterationPerformance} shows the accuracies of \bl{CPJ} when different numbers of feed-forward and back-propagation iterations are reached in the training stage. From \figref{fig:iterationPerformance}, we can see that the prediction accuracy of \bl{CPJ} reaches up to 88.8\% when 10,000 iterations are reached. After that, it stays almost stable even with more iterations. Therefore, we use 10,000 iterations in training \bl{CPJ}.

\myPara{Performance on new datasets}. In the second experiment, we train \bl{CPJ} with all crowdsourced data obtained on the images from any two of the three datasets (\ie, \bl{Toronto}, \bl{MIT} and \bl{ImgSal}) and test the learned metric on the images of the remaining dataset. The main objective is to validate the effectiveness of \bl{CPJ} on new datasets constructed using different configurations in eye-tracking experiments. The performance of \bl{CPJ} and the ten representative metrics $\phi_{1}-\phi_{10}$ are shown in Table~\ref{tab:onNewDataset}.

\begin{table*}
\caption{Accuracies (\%) of ten representative metrics and \bl{CPJ}.}
\centering {
\setlength{\tabcolsep}{3pt}
\begin{tabular}{cccccccccccc}
\toprule	
Metrics      & \bl{AUC} ($\phi_{1}$) & \bl{sAUC} ($\phi_{2}$) & \bl{rAUC} ($\phi_{3}$) & \bl{PRE} ($\phi_{4}$) & \bl{NSS} ($\phi_{5}$) & \bl{SIM} ($\phi_{6}$) & \bl{CC} ($\phi_{7}$) & \bl{IG} ($\phi_{8}$) & \bl{KLD} ($\phi_{9}$) & \bl{EMD} ($\phi_{10}$) &  \bl{CPJ}  \\
\midrule
MT\_I         & 45.8 & 75.3 & 68.6 & 54.1 & 50.7 & 41.1 & 51.1 & 46.8 & 34.0 & 38.6 & \bl{86.0}  \\
IM\_T         & 60.5 & 77.4 & 71.1 & 65.8 & 62.9 & 56.8 & 60.8 & 56.9 & 51.2 & 52.8 & \bl{85.5}  \\
IT\_M         & 57.0 & 71.2 & 68.4 & 63.6 & 58.0 & 50.7 & 57.3 & 55.6 & 48.0 & 48.4 & \bl{85.3}  \\
\bottomrule
\end{tabular}
\begin{tablenotes}
\footnotesize
\item[1] $^*$ MT\_I means \bl{CPJ} is trained on the 300 images from  \bl{MIT} and \bl{Toronto} and tested on the 100 images from \bl{ImgSal}. IM\_T means \bl{CPJ} is trained on the 280 images from \bl{ImgSal} and \bl{MIT} and tested on the 120 images from \bl{Toronto}. IT\_M means \bl{CPJ} is trained on the 220 images from  \bl{ImgSal} and \bl{Toronto} and tested on the 180 images from \bl{MIT}.
\end{tablenotes}
}\label{tab:onNewDataset}
\end{table*}

From Table~\ref{tab:onNewDataset}, we find that \bl{CPJ} still performs the best accuracy in all three cases and outperforms the metrics $\phi_{1}-\phi_{10}$ by at least 8.1\% in IM\_T, even up to 14.1\% in IT\_M. Interestingly, under these three different training and testing settings, the performance ranks of existing metrics is basically the same as that of Table~\ref{tab:onNewImage}. Such consistency in performance ranks and low accuracy scores of representative metrics on different datasets further proves that the existing metrics still lack certain human visual cognition consistency. On the contrary, \bl{CPJ} outperforms the best representative metric on datasets with different eye-tracking configurations. As the performance scores and ranks of \bl{CPJ} stay much more stable across different datasets, it has better capability to assess visual saliency models than the ten representative metrics.

\myPara{Performance on new models}. The third experiment validates whether \bl{CPJ} trained on the crowdsourced data obtained in comparing some specific models can be applied to the assessment of new models. In this experiment, we adopt a leave-one-out strategy in training and testing \bl{CPJ}. That is, we remove one of the three representative models (\ie, {BMS, HFT and SP) from the training stage and use only the instances obtained from the subjective comparisons between the remaining six models for training \bl{CPJ} (about 71.4\% instances). The trained metric is then tested on the instances obtained from subjective comparisons between the removed model and the six models (about 28.6\% instances). The experiment is repeated three times and the results are shown in \tabref{tab:onNewModel}.

\begin{table*}
\caption{Accuracies (\%) of ten representative metrics and \bl{CPJ} when different models are excluded from the training stage.}
\centering {
\setlength{\tabcolsep}{3pt}
\begin{tabular}{cccccccccccc}
\toprule	
Metrics      & \bl{AUC} ($\phi_{1}$) & \bl{sAUC} ($\phi_{2}$) & \bl{rAUC} ($\phi_{3}$) & \bl{PRE} ($\phi_{4}$) & \bl{NSS} ($\phi_{5}$) & \bl{SIM} ($\phi_{6}$) & \bl{CC} ($\phi_{7}$) & \bl{IG} ($\phi_{8}$) & \bl{KLD} ($\phi_{9}$) & \bl{EMD} ($\phi_{10}$) &  \bl{CPJ}  \\
\midrule
%AVG  & 38.9 & 85.0 & 73.6 & 69.1 & 45.3 & 33.0 & 43.1 & 43.5 & 24.3 & 25.2 & 87.3  \\
%IT   & 58.1 & 69.4 & 66.3 & 63.1 & 60.0 & 57.2 & 59.7 & 58.2 & 54.5 & 55.8 & 79.9  \\
%GB   & 55.3 & 67.7 & 65.5 & 60.0 & 58.9 & 53.4 & 58.7 & 55.6 & 50.6 & 53.8 & 78.4  \\
%CA   & 58.7 & 70.5 & 66.7 & 61.8 & 60.3 & 54.4 & 59.7 & 56.3 & 50.0 & 51.4 & 80.2  \\
BMS  & 59.7 & 76.9 & 72.5 & 64.5 & 62.3 & 49.8 & 60.5 & 55.2 & 44.9 & 45.9 & \bl{84.4}  \\
HFT  & 55.1 & 69.5 & 66.4 & 56.1 & 56.6 & 45.2 & 55.5 & 50.2 & 43.3 & 46.7 & \bl{78.4}  \\
SP   & 45.9 & 70.8 & 63.6 & 46.0 & 46.0 & 38.2 & 45.8 & 42.2 & 32.5 & 38.6 & \bl{87.0}  \\
\bottomrule
\end{tabular}
\begin{tablenotes}
\footnotesize
\item[1] $^*$ In each row, instances involving the model in the first column are used for testing, while the remaining ones are used to train \bl{CPJ}.
\end{tablenotes}
}\label{tab:onNewModel}
\end{table*}

From \tabref{tab:onNewModel}, we find that \bl{CPJ} still outperforms the ten representative metrics in comparing a new model with existing ones. Given the new model BMS, HFT or SP, \bl{CPJ} outperforms the best representative metric (\ie, \bl{sAUC} by 7.5\%, 8.5\% and 16.2\%, respectively. As a result, we can safely assume that \bl{CPJ} can be applied to the assessment of newly developed bottom-up (\eg, BMS), spectral (\eg, HFT) and learning-based (\eg, SP) saliency models.

\myPara{Performance on synthetic data}. In the last experiment, we validate the rationality of \bl{CPJ} in assessing saliency maps. We re-train \bl{CPJ} on all the crowdsourced perceptual judgements obtained on the 400 images from all the three datasets, and test the metric on the same number of synthesized data (\ie, $\left\{\left(G_{k},A_{k},G_k\right)|k\in\mb{I}\right\}$ and $\left\{\left(G_{k},B_{k},G_k\right)|k\in\mb{I}\right\}$). Intuitively, the GSM $G_k$ should always outperform better than either of ESM $A_k$ and $B_k$ in subjective tests. The objective of this experiment is to see whether \bl{CPJ} can perfectly capture this attribute, and we find that accuracy of \bl{CPJ} reaches 99.6\%.

Beyond comparing with the GSM that can be viewed as an ``upper bound'', we also test \bl{CPJ} on synthesized data $\left\{\left(A_{k},R_{k},G_k\right)|k\in\mb{I}\right\}$ and $\left\{\left(B_{k},R_{k},G_k\right)|k\in\mb{I}\right\}$. Similarly, comparing with the ``lower-bound'' $R_k$, either of ESM $A_k$ and $B_k$ in subjective tests should always be ``good''. In this case, \bl{CPJ} also achieves an accuracy of 94.3\%. These results ensure that the fixation density maps always achieve the best performance and the random predictions always perform the worst, even though such synthesized data are not used during training \bl{CPJ}.

%In addition, except for using random and ground-truth maps to build such experiments, we also design another baseline $\left\{\left(A_{k},X_{k},G_k\right)|k\in\mb{I}\right\}$ to compare the subjective tests, where $X_{k}$ is a saliency map of the same model with $A_{k}$ from other images. The idea of this baseline is to verify whether high level statistics of saliency maps look similar with the main difference being image-specific is still captured. Here, \bl{CPJ} obtains an accuracy of 73.9\%.

\myPara{Discussion}. From these four experiments, we find that \bl{CPJ} is effective in assessing saliency maps on new images, new datasets, new models and synthetic data. Over different experimental settings, \bl{CPJ} outperforms the ten representative metrics by learning from a variety of subjects assessments. Actually, although different subjects may have different biases in visually comparing the saliency distributions of ESMs and GSMs, the CNN-based framework can automatically infer the most commonly adopted factors shared by various subjects in assessing saliency maps. Therefore, \bl{CPJ} can be viewed as a crowdsourced metric that performs consistently with most of the 16 subjects. Due to this characteristic of the human perception of visual similarity, \bl{CPJ} can evaluate the saliency models from another perspective, making the learned metric a good complement to the existing metrics.

\section{Discussion and Conclusion}\label{sec:conclusion}

In visual saliency estimation, hundreds of models have been proposed to reveal certain characteristics of visual saliency under different assumptions and definitions. Therefore various evaluation metrics are utilized to simultaneously assess the performance of saliency models from multiple perspectives. Usually, the existing metrics are designed to quantitatively compute the performance score of each model without considering perceptual factors. Inspired by the fact that most saliency papers provide representative ESMs and GSMs for qualitative comparisons, the human perception of visual similarity underlying direct visual comparison is very helpful to design new metrics that assess saliency maps as the humans do.

To investigate the key factors that influence the human perception of visual similarity in comparing saliency maps, we conduct extensive subjective tests to find out how saliency maps are assessed by subjects. Besides, to analyze the existing metrics, we propose to quantize the performance of evaluation metrics from the crowdsourced data collected from 16 subjects. A latent assumption here is that even though a subject may have certain biases in assessing saliency maps and models, such biases can be greatly alleviated by summing up the annotations from multiple subjects. Given the crowdsourced data, we find that the top metric that perform the most consistently with the humans only reaches an accuracy of 72.8\% in comparing all the ESM pairs. This indicates that there still exists a large gap between the existing models and models in terms of saliency prediction. % and  from the perspective of the human perception of visual similarity.

To obtain a metric that contains the characteristic of the human perception, we propose to learn a saliency evaluation metric using crowdsourced perceptual judgements based on a two-stream convolutional neural network. Similar to existing saliency evaluation metrics, the learned metric assigns the ESM from an ESM pair a numerical performance score and can capture the key factors shared by various subjects in assessing saliency maps and models. Experimental results show that such a CNN-based metric performs more consistently with human perception of saliency maps and could be a good complement to existing metrics for model comparison and development. In the future work, we will incorporate eye-trackers to study the latent mechanisms by which humans assessing saliency maps. We will also explore the feasibility of building new saliency models under the guidance of our CNN-based metric.

%\section*{Acknowledgments}

%This work was supported in part by grants from the Chinese National Natural Science Foundation under contracts No. 61370113 and No. 61532003, and the Fundamental Research Funds for the Central Universities.

\bibliographystyle{IEEEtran}
\bibliography{RefSal-20160405}

%\vspace{-2in}

\begin{biography}[{\includegraphics[width=1in,height=1.25in,clip,keepaspectratio]{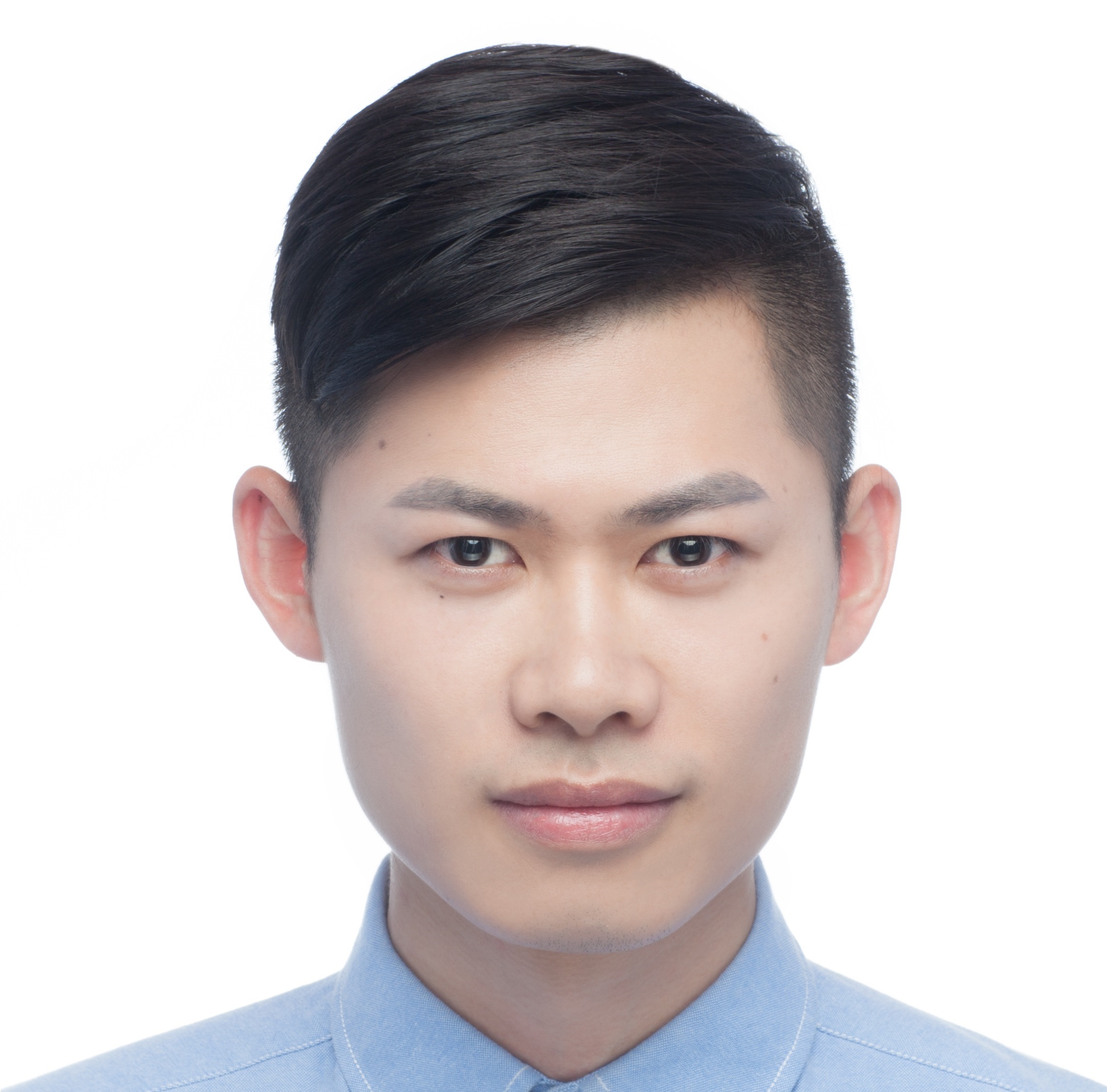}}]{Changqun Xia} is currently pursuing the Ph.D. degree with the State Key Laboratory of Virtual Reality Technology and System, School of Computer Science and Engineering, Beihang University. His research interests include computer vision and image understanding.
\end{biography}

%\vspace{-2in}

\begin{biography}[{\includegraphics[width=1in,height=1.25in,clip,keepaspectratio]{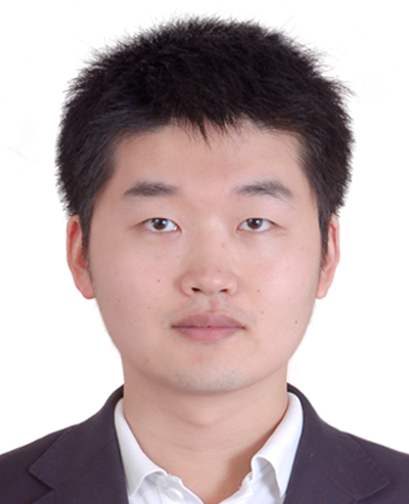}}]{Jia Li} is currently an associate Professor with the School of Computer Science and Engineering, Beihang University, Beijing, China. He received the B.E. degree from Tsinghua University in 2005 and the Ph.D. degree from the Institute of Computing Technology, Chinese Academy of Sciences, in 2011. Before he joined Beihang University in Jun. 2014, he servered as a researcher in several multimedia groups of Nanyang Technological University, Peking University and Shanda Innovations. He is the author or coauthor of over 50 technical articles in refereed journals and conferences such as TPAMI, IJCV, TIP, CVPR, ICCV and ACM MM. His research interests include computer vision and multimedia big data, especially the learning-based visual content understanding. He is a senior member of IEEE and CCF.
\end{biography}

%\vspace{-2in}

\begin{biography}[{\includegraphics[width=1in,height=1.25in,clip,keepaspectratio]{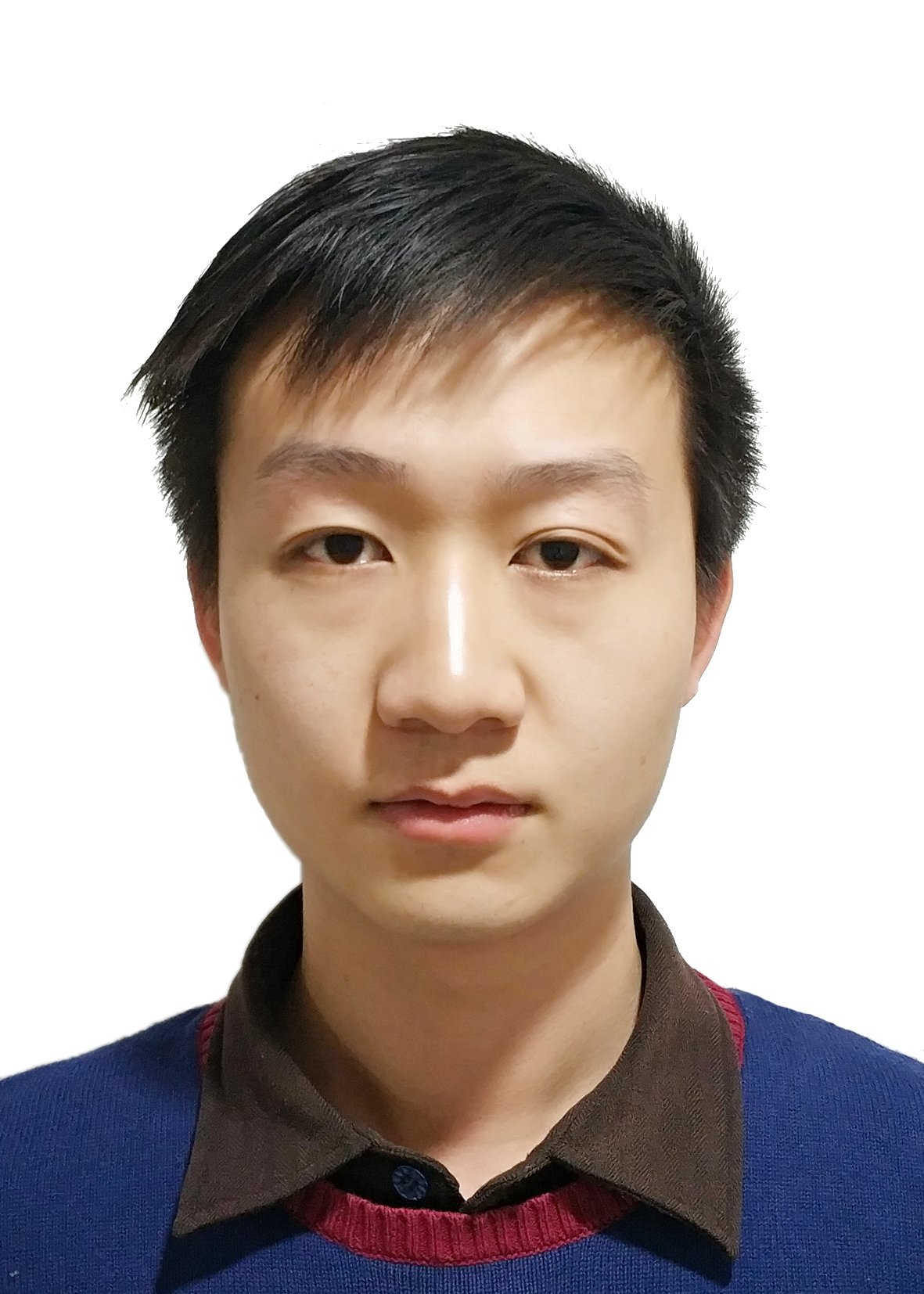}}]{Jinming Su} is currently pursuing the Master degress with the State Key Laboratory of Virtual Reality Technology and Systems, School of Computer Science and Engineering, Beihang University. His research interests include computer vision and machine learning.
\end{biography}

%\vspace{-2in}

\begin{biography}[{\includegraphics[width=1in,height=1.25in,clip,keepaspectratio]{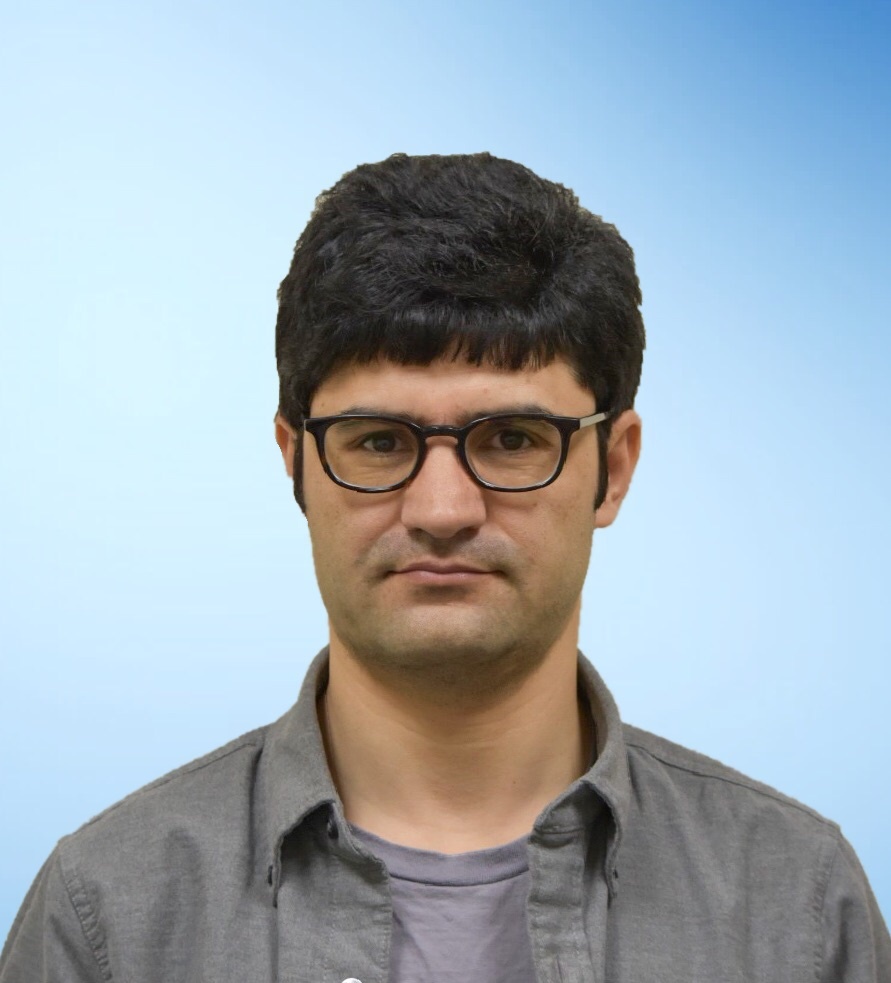}}]{Ali Borji} Ali Borji received his BS and MS degrees
in computer engineering from Petroleum University of Technology, Tehran, Iran, 2001 and Shiraz University, Shiraz, Iran, 2004, respectively. He did his Ph.D. in cognitive neurosciences at Institute for Studies in Fundamental Sciences (IPM) in Tehran, Iran, 2009 and spent four years as a postdoctoral scholar at iLab, University of Southern California from 2010 to 2014. He is currently an assistant professor at the Center for Research in Computer Vision and the Department of Computer Science at the University of Central Florida. His research interests include visual attention, active learning, object and scene recognition, and cognitive and computational neurosciences.
\end{biography}

\end{document}